\documentclass[conference]{IEEEtran}

\include{preamble}

\usepackage{comment}

\usepackage{enumitem}
\usepackage{titlesec}
\usepackage{booktabs}
\usepackage{tabularx}
\usepackage{graphicx} 
\usepackage{algorithm}
\usepackage{algpseudocode}
\usepackage{amsmath, amssymb}
\usepackage{subcaption}
\usepackage{xcolor}
\usepackage{multirow}
\usepackage[colorlinks=true, citecolor=blue, linkcolor=blue, urlcolor=blue]{hyperref}

\usepackage{xcolor}
\usepackage[a4paper, total={184mm,239mm}]{geometry}
\def\BibTeX{{\rm B\kern-.05em{\sc i\kern-.025em b}\kern-.08em
    T\kern-.1667em\lower.7ex\hbox{E}\kern-.125emX}}

\begin{document}

\title{RocketPPA: Code-Level Power, Performance, and Area Prediction via LLM and Mixture of Experts
}

\author{
  \IEEEauthorblockN{Armin Abdollahi, Mehdi Kamal, and Massoud Pedram}
  \IEEEauthorblockA{Department of Electrical and Computer Engineering,
    University of Southern California, Los Angeles, CA, USA\\
    \{arminabd, mehdi.kamal, pedram\}@usc.edu}
}

\maketitle


\begin{abstract}
This paper presents RocketPPA, a novel ultra-fast power, performance (delay), and area (PPA) estimator operating directly at the code-level abstraction using HDL code as input. The key technical innovation is its LLM-based regression model, which uniquely integrates a large language model (LLM) with a mixture-of-experts (MoE) architecture composed of multilayer perceptrons (MLPs). The LLM interprets the input HDL code and then utilizes its final hidden-layer representations to predict PPA metrics. Low-rank adaptation (LoRA) is used for parameter-efficient fine-tuning to enable efficient LLM training. Furthermore, the work includes the development of an LLM-based HDL code repair framework to generate a large and synthesizable training dataset. Experimental results on the VerilogEval benchmark demonstrate that RocketPPA achieves significant improvements in the accuracy of PPA estimation compared to previous state-of-the-art methods like Llama3-MetRex-8B. Specifically, at a 10\% relative error threshold, RocketPPA enhances the pass rate for area prediction by 13.6\%, delay by 9.4\%, and power by 14.7\%. At a 20\% threshold, the improvements are 9.6\% for area, 10.8\% for delay, and 18.5\% for power. Moreover, RocketPPA achieves a speedup of over 20× compared to MetRex and 30× over MasterRTL in processing the test set. The impact of RocketPPA is the potential to substantially accelerate the hardware design process by providing accurate PPA estimations early in the design cycle, thus avoiding the overhead of manual feature engineering and time-consuming synthesis flows.



\end{abstract}

\IEEEpeerreviewmaketitle
\vspace{-5pt}
\section{Introduction}
PPA estimators are at the heart of modern VLSI design, as they critically influence the functionality and efficiency of hardware implementations \cite{ref101}. In conventional electronic design automation (EDA), high-level approaches to PPA estimation typically involve abstract modeling techniques, simulation-based analyses, and synthesis optimizations that produce approximate metrics \cite{ref102}. Although these traditional methods provide a helpful baseline, they often fail to capture nuanced interactions in increasingly complex circuits \cite{ref103}. Their limitations include significant computational overhead, the reliance on manual calibration, and a lack of granularity that can impede the ability to fine-tune designs for optimal power, delay, and area performance \cite{ref104}.
These inherent limitations in traditional PPA estimation methods have started the exploration of alternative approaches that promise to capture the complexities of modern VLSI design more effectively.

In this context, large language models (LLMs) have recently demonstrated considerable potential in automating hardware design workflows, particularly for Verilog code generation and analysis. Starting with solutions such as BetterV \cite{ref1} and ChipNeMo \cite{ref2}, researchers have shown that domain-specific fine-tuning can allow LLMs to generate high-quality Verilog modules that meet syntactic and even partial functional requirements. Building on these early results, multi-expert modeling paradigms like \cite{ref3} and hierarchical or specialized data-curation pipelines like \cite{ref4, ref12} have significantly expanded LLM performance, making it possible to target more complex scenarios in electronic design automation. Meanwhile, others have broadened the scope to include Verilog understanding \cite{ref5}, hallucination mitigation \cite{ref6}, hierarchical frameworks \cite{ref7}, and open-source linting \cite{ref8}. These advancements underscore the dramatic shift in chip design and verification by LLMs, creating a new research trajectory where code generation, debugging, and performance tuning converge.

In this work, we propose an LLM-based regression model for estimating the design metrics of a given HDL Verilog code. The model accelerates the hardware design process by directly predicting PPA metrics at the code level.
Our approach utilizes an LLM to analyze and understand the input HDL code, generating a latent representation based on its final hidden layer results. This representation is shared across separate MoE MLPs, each dedicated to predicting a specific design metric.
We employ a normalization strategy to address the large dynamic range of metric values that improves the accuracy of the model prediction.
To improve training efficiency and allow flexibility in using different foundation models, we apply LoRA to fine-tune the LLM. 
For training the proposed LLM-based PPA estimator, we generate a large set of syntactically correct and synthesizable HDL designs and train the model on them. 
Additionally, we evaluate its efficiency by comparing its estimations with those of two state-of-the-art (SOTA) code-level PPA estimators.




Based on the above explanations, we can summarize the contributions of this work as follows.

\begin{itemize}
    \item Introducing an end-to-end LLM-based neural architecture that directly translates Verilog code into power, delay, and area metrics without additional processing steps.
    \item Integrating a mixture-of-experts extension to the model's regression part, enhancing the design parameters' prediction.
    \item Proposing a normalizing approach to improve the accuracy of the model. 
    
    
\end{itemize}

This paper is organized as follows. Section II reviews some previous work. The details of the proposed LLM-based PPA estimator are discussed in Section III. Section IV evaluates the efficacy of the RocketPPA compared to earlier works. Finally, the paper is concluded in Section V.

\vspace{-5pt}
\section{Related Work}
In this section, we review prior work on the use of LLMs in hardware design.
Early efforts to harness LLMs to generate valid Verilog modules include BetterV \cite{ref1}, which introduced discriminative guidance, and ChipNeMo \cite{ref2}, which pioneered domain-adaptive tokenization and pre-training. Following these, multi-expert architectures \cite{ref3} and hallucination-mitigation strategies \cite{ref6} extended generation accuracy by focusing on syntactic correctness and domain alignment. Other works have explored hierarchical decomposition \cite{ref7}, prompting frameworks \cite{ref13}, and multi-agent ecosystems \cite{ref10} to enhance code quality. Another crucial direction involves cleaning up and augmenting Verilog datasets for improved model training. For example, CRAFTRTL \cite{ref4} introduced correct-by-construction non-textual data for code generation models, while OriGen \cite{ref12} leveraged code-to-code augmentation and self-reflection to refine Verilog sets further. These studies have significantly reduced minor syntax errors by synthesizing large volumes of high-fidelity data, enabling more robust fine-tuning.

Beyond raw generation, researchers have begun to quantify how well LLMs understand Verilog semantics. DeepRTL \cite{ref5} presents a unified representation model that supports both generation and comprehension, and LintLLM \cite{ref8} proposes a pipeline for automated defect detection. For more targeted benchmarks, RTLLM \cite{ref15} and VerilogEval \cite{ref18} offer tasks and datasets to measure functional correctness. In contrast, MetRex \cite{ref11}  specifically addressed the post-synthesis metrics (area, delay, power). MetRex~\cite{ref11}\enspace provides a large and automatically cleaned corpus of synthesized Verilog designs paired with post‑synthesis area, delay, and static power reports. A chain‑of‑thought template guides LLMs through gate counts and critical path reasoning, and supervised LoRA fine‑tuning on this dataset demonstrates that metric estimation can be performed directly from HDL text without feature extraction.
This ongoing shift toward PPA-aware generation is underscored by advanced frameworks such as \cite{ref19, ref20}, which leverage multi-stage prompts and search strategies to generate code meeting power or timing goals.

As these models become fully integrated into the chip design workflow, researchers have also looked at watermarking \cite{ref16} and IP protection. Frameworks like \cite{ref21} highlight security and trust concerns with LLM-based EDA, advocating stronger safeguards against malicious or unauthorized code usage. Meanwhile, solutions like OPL4GPT \cite{ref22} explore how alternate programming paradigms (e.g., C++-to-hardware flows) compare to conventional Verilog flows for hardware acceleration. The field has rapidly evolved from simple Verilog text generation to comprehensive frameworks integrating domain-adaptive data curation, multi-expert architectures, and advanced debugging. Our work builds on these efforts by presenting a fine‑tuning strategy (via LoRA and an MLP-based mixture-of-experts) to address performance estimation in hardware design jointly. We aim to make LLM-based solutions more practically viable for next-generation EDA, moving beyond code syntax to incorporate critical metrics that drive real-world chip design decisions.

\section{Proposed LLM-based PPA-Estimator Method}

\subsection{Model Architecture}
\begin{figure*}[!t]
\centering
\includegraphics[width=5.5in]{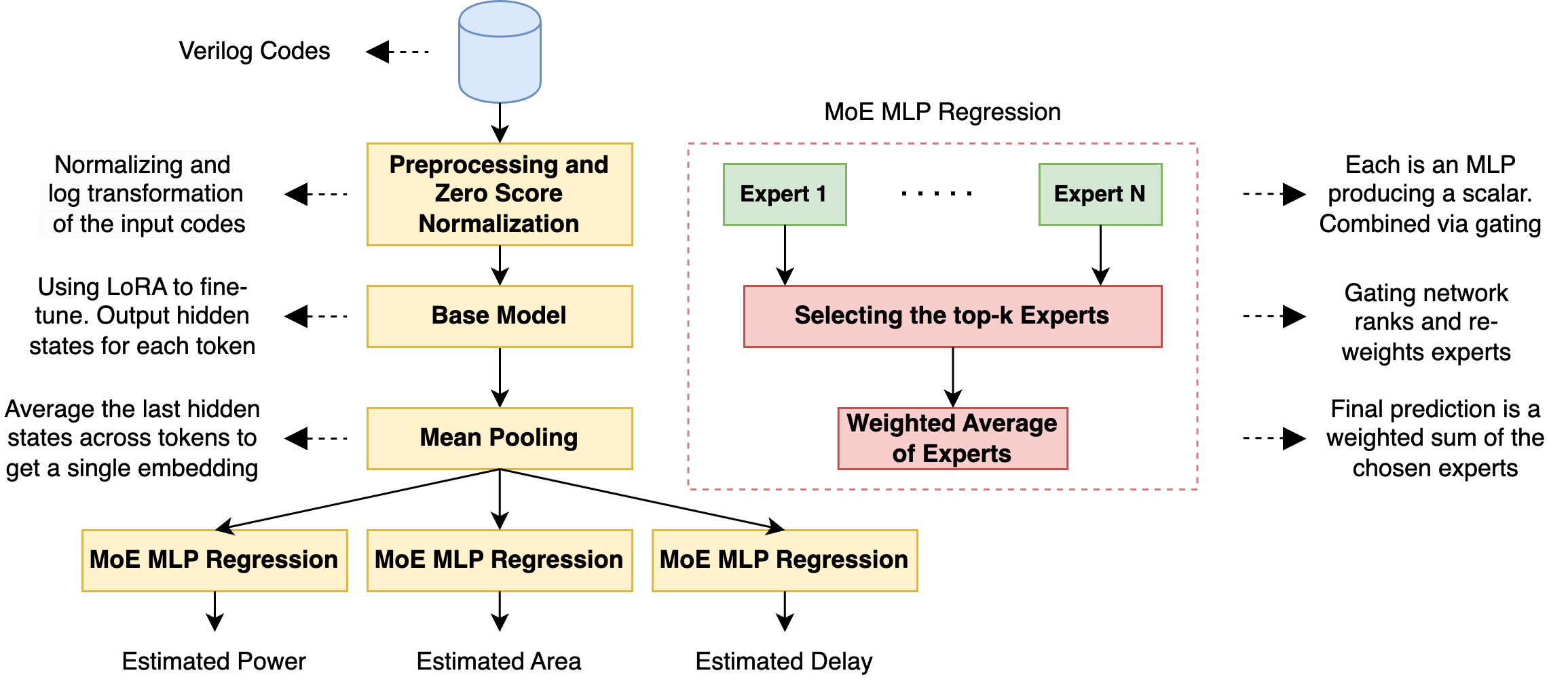}
\caption{\small The model architecture of the proposed PPA estimator.}
\label{fig:Arch2}
\end{figure*}

Figure~\ref{fig:Arch2} shows the general architecture in which the proposed neural architecture consists of three main components: (1) a foundational LLM for the extraction of features, (2) an MoE regression neural model for the prediction of the final metric, and (3) a LoRA‐based parameter-efficient fine-tuning scheme that adjusts the parameters of the foundation model.

We employ a base model as our backbone to encode Verilog code inputs into high-dimensional representations. Unlike the causal language model heads commonly used for text generation, we switch to a regression-friendly interface. The model produces a stack of hidden states $\{h^1, h^2, \dots, h^L\}$ during the forward pass. We take the last hidden state $h^L$ and perform a simple average pooling along the sequence dimension defined in Equation~\ref{eq:pooled_output}, where $n$ denotes the number of tokens in the code. This pooled output (which has, for example, a dimension of 4096 in CodeLlama-7B) then serves as input to our Mixture of Experts regressor.

\begin{equation}
O_{pooled} = \frac{1}{n} \sum_{i=1}^{n} h_i^L
\label{eq:pooled_output}
\end{equation}

To capture the diversity of hardware modules and thereby avoid biasing a single dense network toward any single design regime, we incorporate a MoE technique. We define $N$ distinct expert subnetworks, each comprising multiple fully connected layers with activations and dropout. Each expert is precisely implemented as an MLP that produces an individual scalar prediction $y_k$ for the target performance metric. A small linear layer maps the pooled output from LLM to a distribution over the $N$ experts. Formally, given the pooled output, the gating network computes the gating weights by:

\begin{equation}
\alpha = \text{Softmax}(W_{\text{gate}} \times O_{pooled})
\label{eq:alpha}
\end{equation}

\noindent where $\alpha \in \mathbb{R}^N$ and each element $\alpha_k$ represents the initial weight assigned to the $k^{th}$ expert. To focus the model's capacity on the most relevant experts, we employ a top‑k gating mechanism. Let $K$ be the set of indices corresponding to the top‑k experts with the highest values in $\alpha$. We then renormalize the gating weights for these experts by:

\begin{equation}
\tilde{\alpha}_k = \frac{\alpha_k}{\sum_{j \in K} \alpha_j}, \quad \text{for } k \in K.
\label{eq:alpha_normalization}
\end{equation}

This normalization ensures that the weights of the selected experts sum up to 1, effectively concentrating the predictive capacity of the model on the most pertinent outputs of the experts. Finally, the overall prediction $\hat{y}$ is calculated as a weighted sum of the outputs of only the top-k experts:

\begin{equation}
\hat{y} = \sum_{k=1}^{N} \tilde{\alpha}_k y_k
\label{eq:y_hat}
\end{equation}

Thus, the gating network first projects the (e.g., 4096-dimensional) pooled representation of the model into an N-dimensional space, yielding a probability distribution over the experts via a softmax operation. The top‑k experts are then selected, and their weights are renormalized to form a focused aggregation of expert predictions. This design allows the model to take advantage of the specialized knowledge of a subset of experts most familiar with a given input, thus improving the performance of regression for predicting hardware metrics such as power, delay, and area.

We adopt a parameter-efficient fine-tuning approach using LoRA, which significantly reduces computational overhead by focusing on a subset of parameters rather than fine-tuning all LLM components. This strategy maintains the expressiveness of the model while limiting the number of trainable parameters. In contrast, the MoE regressor is kept fully trainable, allowing its gating network and expert components to effectively adapt to the diverse range of Verilog designs and PPA values in the dataset.

\begin{figure}[!t]
\centering
\includegraphics[width=3.0in]{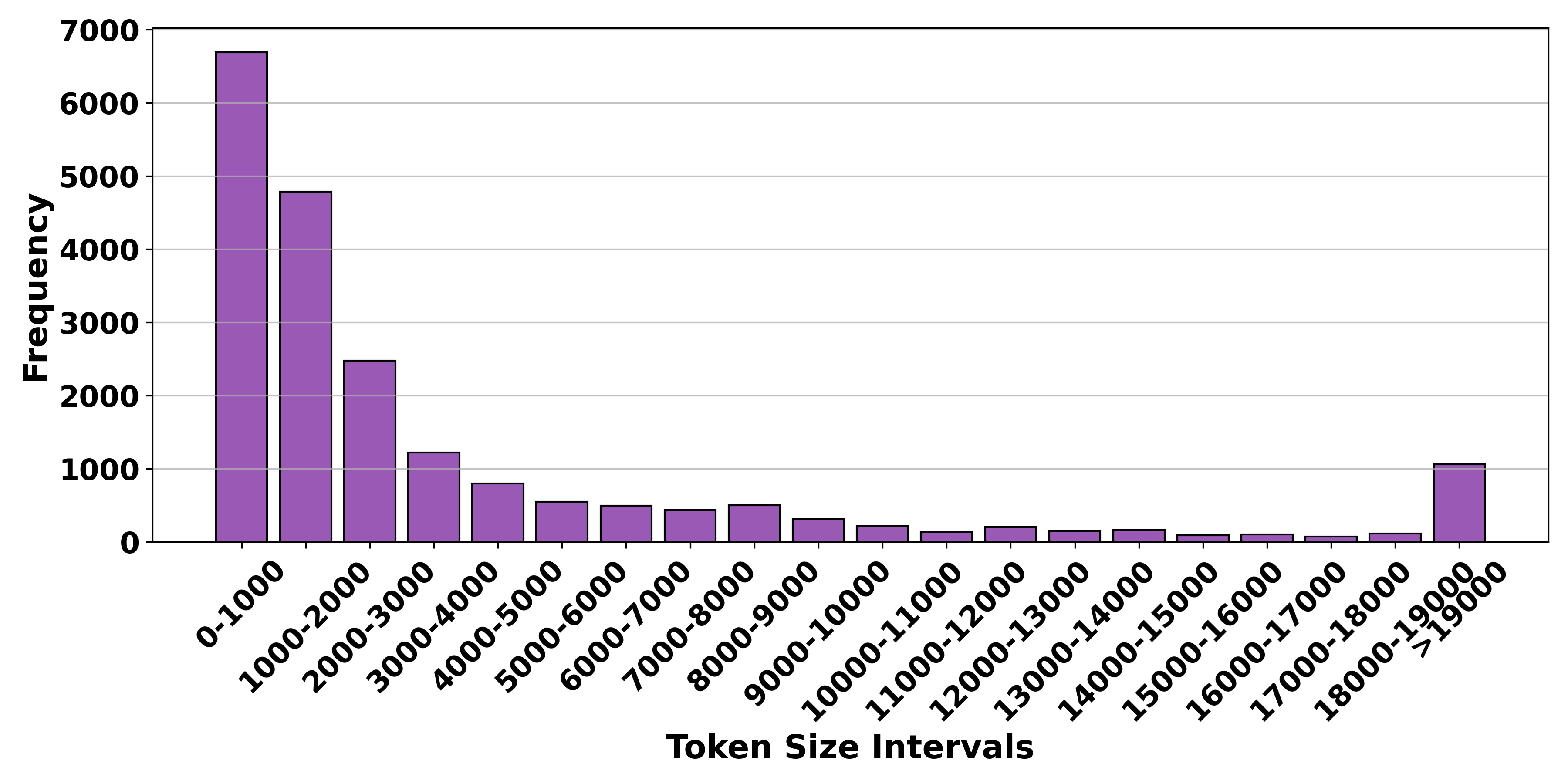}
\caption{\small Token size distribution of the Verilog files in the studied dataset.}
\label{fig:Dist}
\end{figure}

As an example, Figure \ref{fig:Dist} illustrates the token size distribution of Verilog files in the dataset studied in this work. The results show a wide distribution, with some Verilog files exceeding the input token limit of the foundational LLM model employed.
We address the limitation of the LLM's input size (i.e., number of tokens) by splitting each lengthy Verilog file into smaller, non-overlapping, fixed-size token fragments (e.g., each fragment contains 512 tokens). Each fragment is passed individually through the model, producing a separate embedding for that code fragment. These fragment embeddings are subsequently combined, e.g., by taking their average, to create a single, unified embedding representing the entire Verilog file. This approach preserves all code information without truncation because every token belongs to exactly one fragment. Therefore, we maintain the maximum sequence length limit while still capturing the complete context of large Verilog inputs \cite{ref201}.

\begin{figure*}[]
  \centering
  \begin{subfigure}[b]{0.32\textwidth}
    \centering
    \includegraphics[width=\textwidth]{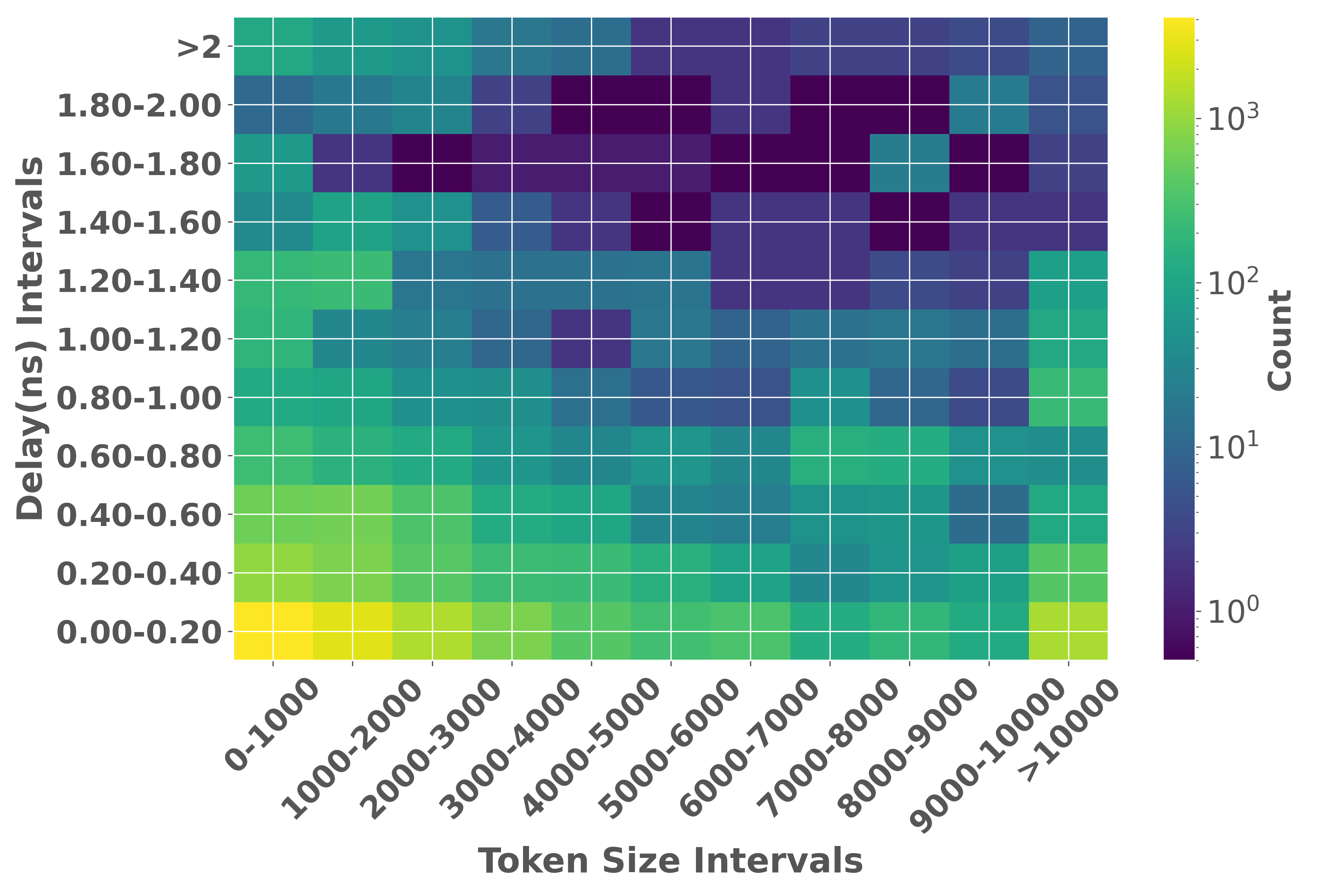}
    \caption{}
    \label{fig:delay2d}
  \end{subfigure}
  \hfill
  \begin{subfigure}[b]{0.32\textwidth}
    \centering
    \includegraphics[width=\textwidth]{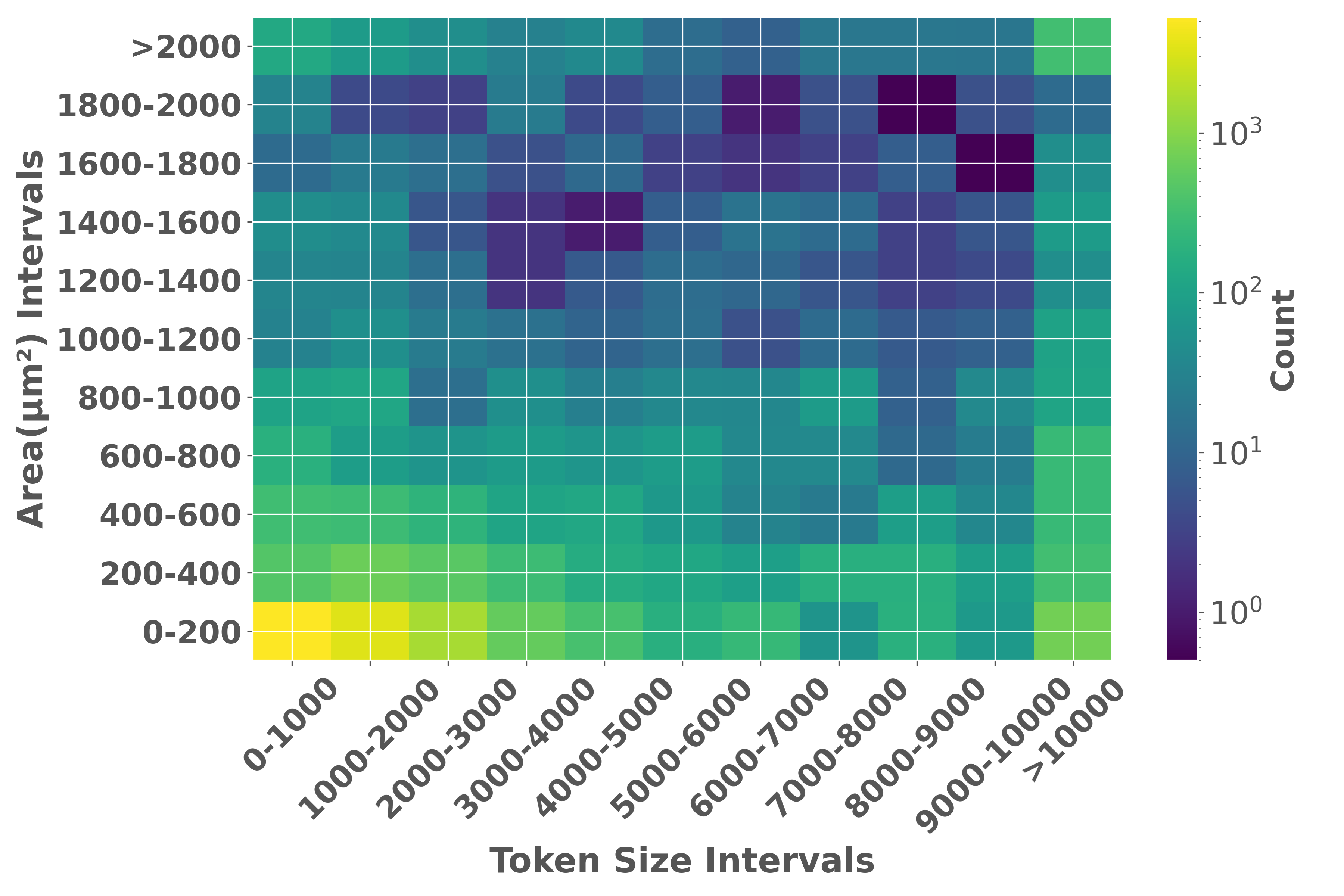}
    \caption{}
    \label{fig:area2d}
  \end{subfigure}
  \hfill
  \begin{subfigure}[b]{0.32\textwidth}
    \centering
    \includegraphics[width=\textwidth]{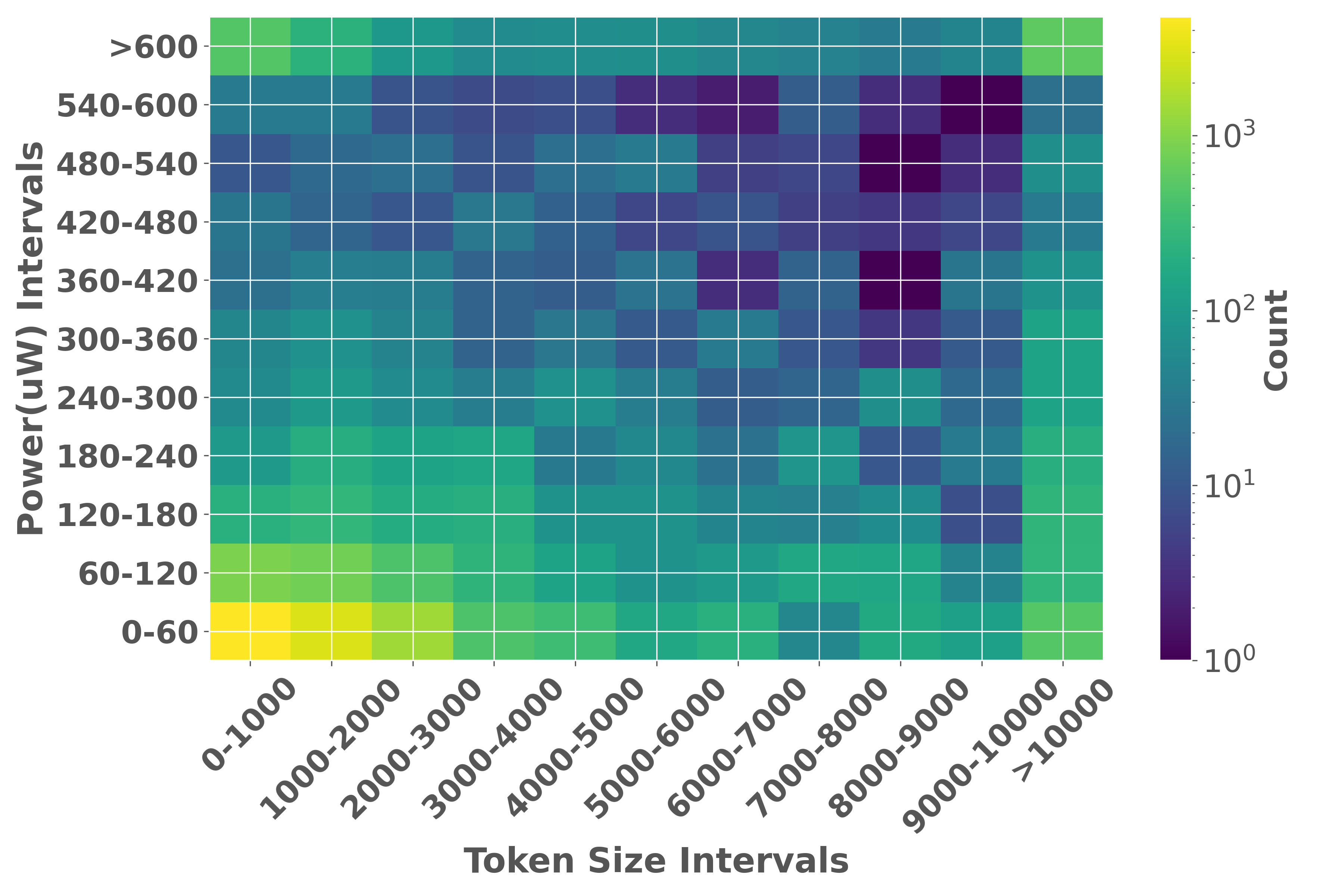}
    \caption{}
    \label{fig:power2d}
  \end{subfigure}
  
  \caption{2D Heatmap of Token Size vs. Power, Delay, and Area}
  \label{fig:2dplots}
\end{figure*}

\subsection{Data Pre-processing and Training flow}

Since raw performance metrics can span several orders of magnitude (see Figure \ref{fig:2dplots}), the log transformation is applied to compress the range and stabilize the numerical values. Specifically, for each performance metric $x_i$, we compute $log(x_i + \epsilon)$ where $\epsilon$ is a small constant to avoid taking the logarithm of zero. This transformation reduces the variance and mitigates the impact of extreme values. To ensure stable training, we normalize the logarithmic transformation of the performance metric values by computing the mean $\mu$ and the standard deviation $\sigma$ over the training set and, subsequently, applying a z-score normalization, which maps the logarithmically transformed values to a standard normal distribution with zero mean and unit variance. Hence, the normalized input performance metric ($\hat{x}_i$) is obtained by:

\begin{equation}
\hat{x}_i = \frac{\log(x_i + \epsilon) - \mu}{\sigma}
\label{eq:normalization}
\end{equation}

For training loss, we use the smooth $\ell_1$ (Huber) loss, which offers a balance between mean square error (MSE) and mean absolute error (MAE). The smooth $\ell_1$ loss behaves quadratically when the prediction error is small, ensuring fine-grained adjustments. However, it transitions to linear behavior when the error is large, thereby reducing the influence of outliers. Hence, we suggest the loss defined by Equation (\ref{eq:smooth_l1}). This piecewise formulation ensures that the loss function remains robust to outliers while providing smooth gradients when errors are small, which is essential for stable convergence.

\begin{equation}
L =
\begin{cases}
0.5\, (\hat{y}_i - z_i)^2, & \text{if } |\hat{y}^i - z_i| < 1, \\
|\hat{y}_i - z_i| - 0.5, & \text{otherwise.}
\end{cases}
\label{eq:smooth_l1}
\end{equation}

In this equation, $\hat{y}_i$ represents the predicted output of the model for the $i^{th}$ Verilog code metric, while $z_i$ denotes its corresponding actual value. Log transformation, z-score normalization, and smooth $\ell_1$ loss work together to handle the wide range of performance metric values. This normalization scheme standardizes the data, ensuring that the regression model learns effectively, while the Smooth $\ell_1$ loss balances sensitivity and robustness during training.

\begin{figure}[]
\centering
\includegraphics[width=2.5in]{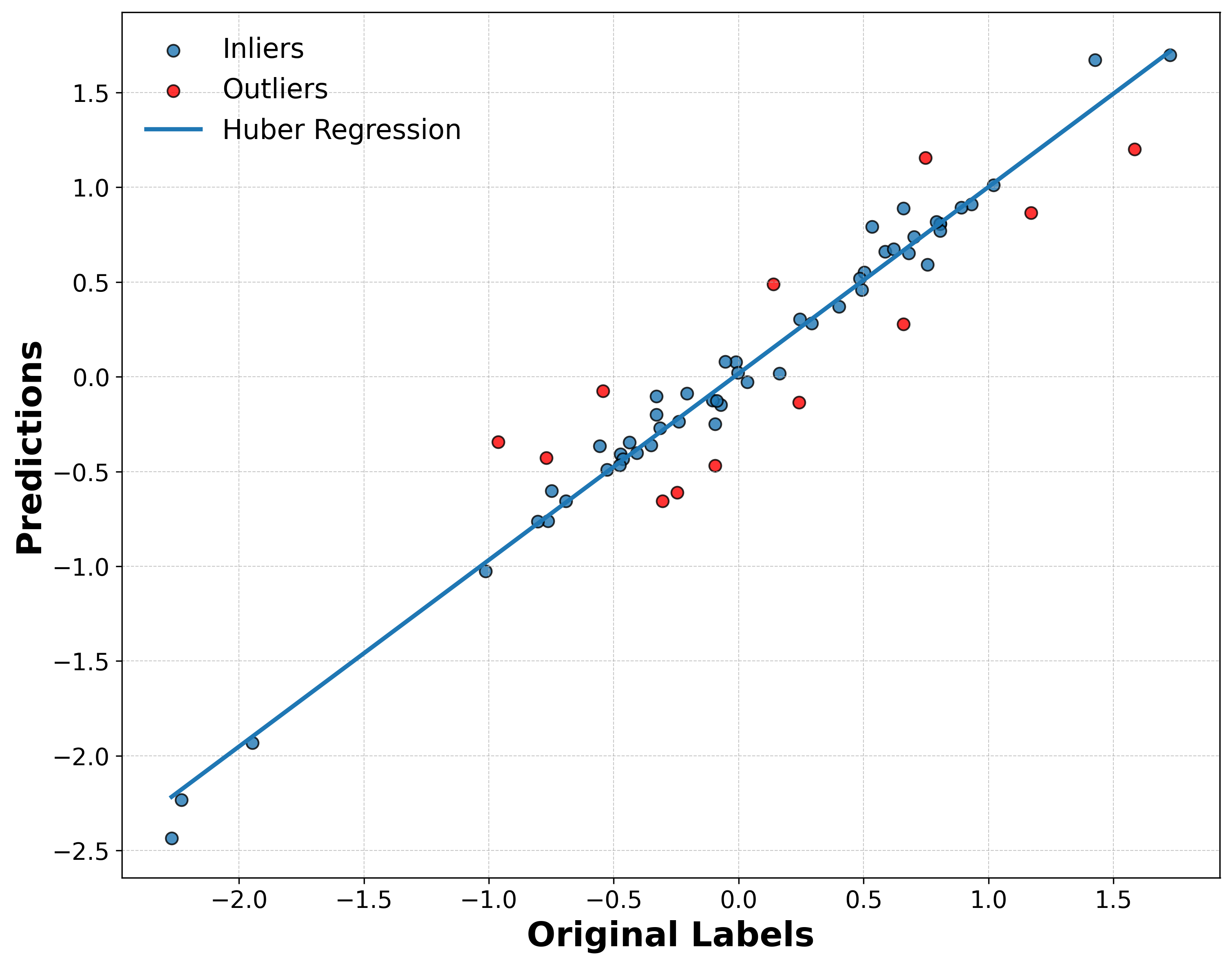}
\caption{\small Hubor loss demonstration on a single batch of data during training}
\label{fig:Hubor}
\end{figure}

Figure~\ref{fig:Hubor} illustrates predictions versus ground-truth labels at a mid-epoch stage of training under the Huber loss. The blue points (inliers) lie within the Huber threshold, where the loss behaves quadratically, much like MSE, so these points strongly influence the regression fit. However, red points (outliers) trigger the Huber loss linear regime, which means that large residuals have a reduced impact on parameter updates. The blue line represents the Huber regression fit that balances these two behaviors: It remains sensitive to most inliers while preventing extreme outliers from dominating the model.

\begin{figure*}[!t]
\centering
\includegraphics[width=6in]{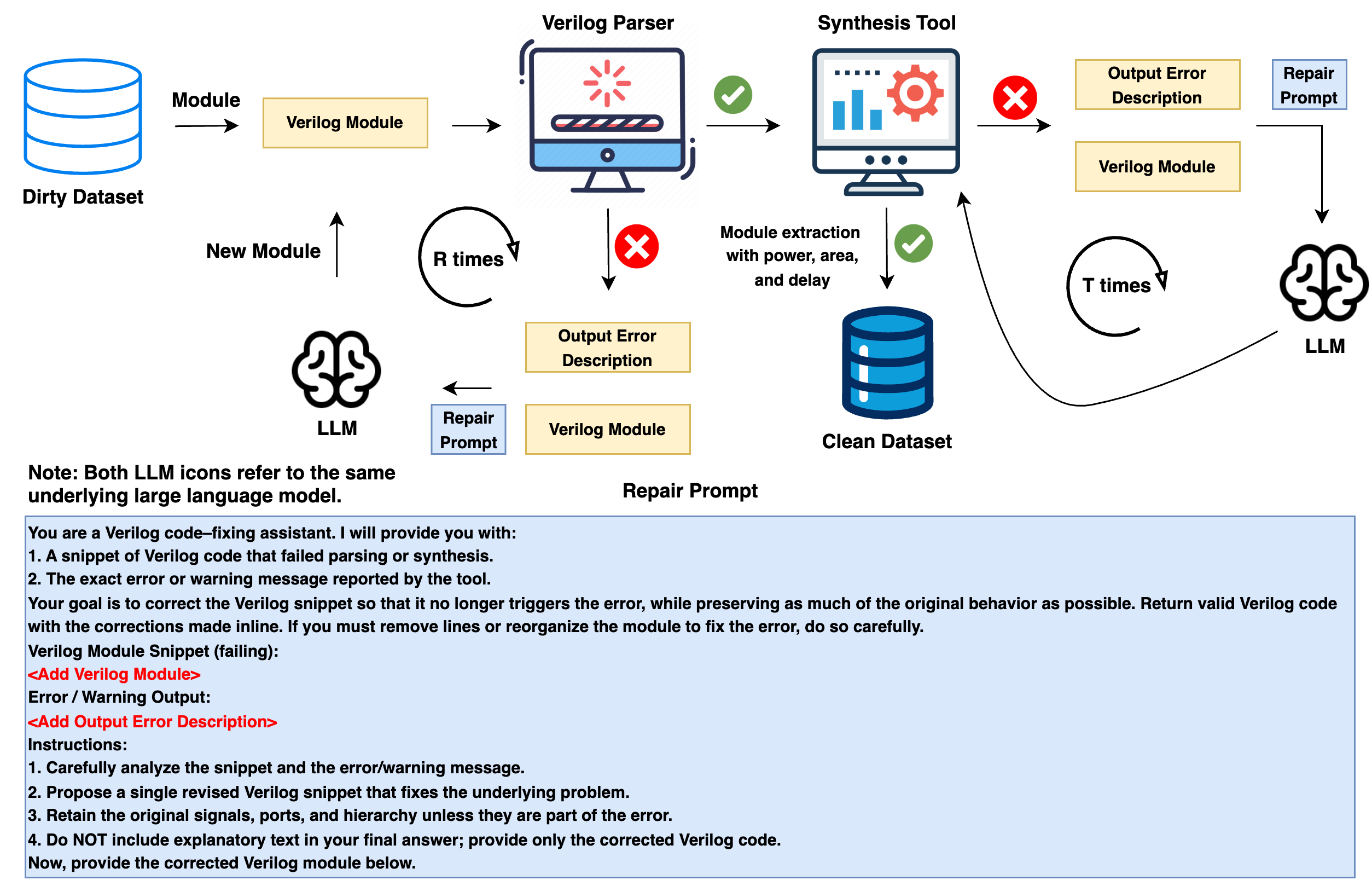}
\caption{The proposed HDL code repair framework. }
\label{fig:Arch1}
\end{figure*}

\section{Results and Discussion}

\subsection{Experimental Setup}
For the evaluation study, the dataset was generated by aggregating Verilog modules (some faulty) from three primary sources to ensure a broad coverage of the design lengths and complexities. Due to the significant number of incorrect codes, we used an internal HDL code repair framework (detailed in Figure~\ref{fig:Arch1}) to correct syntax errors and ensure synthesizability.

Specifically, the generated dataset has 9,239 modules from the MG-Verilog repository, 7,791 modules from a curated GitHub collection (verilog\_github), and 3,923 modules from the VeriGen dataset. These modules span a broad spectrum, from simple combinational circuits to complex finite-state machines, ensuring that our data set is representative of the challenges encountered in practical hardware design. The details of the generated dataset are provided in Table~\ref{tab:dataset_composition_simple}.
We split our purified data into 80-20 training versus validation splits. Each sample consists of a Verilog code snippet and the log-transformed variant of the corresponding performance metric.

For comparative studies, we used sample codes (138 HDL codes) from the VerilogEval benchmark \cite{ref18}. The dataset is divided into three levels. Specifically, we also performed a set of studies on the codes in Level 3. These HDL codes represent the most complex part of the VerilogEval dataset. This level comprises 72 modules (over 50\% of the total of 138) and includes designs such as finite-state machines, complex combinational logic, and advanced sequential circuits. These modules span a wide gate count range, up to 607 gates, and pose a significant challenge for PPA prediction models due to their structural depth and behavioral intricacies. Note that this dataset has been used only for evaluation, not training.

Our training workflow completes in around 11 hours on a single NVIDIA RTX A6000 GPU with 48GB of VRAM for a full 24-epoch schedule. Despite the large parameter count of employed foundation models, our LoRA-based parameter-efficient fine-tuning keeps GPU memory usage at roughly 41GB, comfortably fitting on one GPU card without requiring model parallelism or multi-GPU setups. 

The model was trained 10 times, with different initial random seeds. Hence, each experiment was performed ten times, and the observed standard deviation was below 1\%. This small variability arises primarily from inherent stochastic processes during training (e.g., the randomness introduced by dropout and weight initialization). This indicates that our results are stable under modest changes in initialization and data shuffling. 


\begin{table}[ht]
\centering
\caption{Dataset Composition}
\begin{tabular}{lc}
\hline
Source          & Number of Modules \\ \hline
MG-Verilog      & 9,239            \\
verilog\_github & 7,791             \\
VeriGen         & 3,923             \\ \hline
Total           & 20,953            \\ \hline
\end{tabular}
\label{tab:dataset_composition_simple}
\end{table}

To assess the efficacy of the proposed estimator, we compare its efficiency with those of MetRex~\cite{ref11} (with the Llama3-8B model as the LLM employed) and MasterRTL~\cite{ref23}. Note that MetRex can only estimate static power, in contrast to RocketPPA and MasterRTL, which can evaluate the overall power consumption in addition to the static power.
MasterRTL~\cite{ref23}\enspace tackles the PPA estimation task via a pre‑synthesis route: it converts RTL into a bit‑level Simple Operator Graph (SOG) that aligns one‑to‑one with the synthesized netlist. Separate tree‑based models then predict timing, power, and area from SOG features, offering strong cross‑design generalization but at the cost of a CPU‑intensive graph‑construction pipeline.

The Pass Rate metric (introduced in \cite{ref11}) has been used to evaluate the methods. This metric represents the percentage of HDL code input instances for which the estimator predicts a design metric (area, delay, and power) with an error below a predefined threshold (i.e., $\theta$). In this context, the error refers to the relative error (RE).
Equation~(\ref{eq:pass_rate}) defines the Pass Rate at threshold $\theta$, computed on $M$ samples. Specifically, for each sample $i$, we check whether its relative error $RE_i$ is below or equal to $t\theta\%$. The indicator function $\mathbf{1}$ returns 1 if the condition holds and 0 otherwise, and the pass rate is simply the average of these indicator values in all $M$ samples. 

\begin{equation}
\label{eq:pass_rate}
\text{PassRate}(\theta) \;=\; \frac{1}{M} \sum_{i=1}^{M} \mathbf{1}\!\Bigl(\text{RE}_i \le \theta\Bigr)
\end{equation}

Finally, for the comparative studies in this paper, we employed the LLaMA-3.1-8B-Instruct model \cite{ref120} as the LLM engine of RocketPPA. We also used CodeLlama-7B \cite{ref121} and LLaMA-3-8B \cite{ref120} for ablation studies.
For every design metric (i.e., area, delay, and power dissipation) we considered an MoE head with $N\!=\!6$ experts; during inference, the top‑$k$ gating network kept only the $k\!=\!3$ highest‑scoring experts for the weighted sum. Each expert was a three‑layer MLP, hence the MoE block adds 4.7M parameters. Low‑Rank Adaptation at rank \(r=16\) was applied to each transformer block’s weight matrices, adding approximately \(8.4\)M trainable parameters (about \(0.11\%\), in the case of LLaMA-3.1-8B-Instruct model).

\subsection{Results}
\subsubsection{Comparison Study}

Table~\ref{tab:comparison} compares the pass rate metric across the considered methods for area, delay, total power (static Power + dynamic power), and static power, evaluated under threshold values $\theta$ of 10\% and 20\%. In this study, all HDL codes of VerilogEval were used for evaluation.
As the 10\% threshold, our MoE model achieves an area prediction pass rate of 71.6\%, which represents a 13.6 percentage point improvement over Llama3-MetRex-8B (58.0\%) and a 20.4 percentage point increase over MasterRTL (51.2\%). Our model achieves 57.2\% pass rate for delay estimation, outperforming Llama3-MetRex-8B’s 47.8\% (a 9.4 percentage point gain), and MasterRTL's 50.7\% (a 6.5\% increase). Regarding total power dissipation, our MoE variant reaches 55.0\% pass rate, improving by 9.9\% percentage points over MasterRTL (45.1\%). For Static Power, our proposed method can achieve 56.7\%, 14.7\% higher than Llama3-MetRex-8B (42.0\%), and 9.2\% larger than MasterRTL (47.5\%). At the more lenient 20\% threshold, we observe parallel gains across all metrics: the area‑prediction pass rate climbs to 84.6\% (a 9.6‑percentage‑point improvement over Llama3‑MetRex‑8B’s 75.0\% and a 3.1\% uplift over MasterRTL’s 78.1\%); the delay‑prediction pass rate rises to 74.9\% (10.8 points above Llama3‑MetRex‑8B’s 64.1\% and 7.0\% above MasterRTL); the static‑power pass rate reaches 72.8\% (18.5 points higher than Llama3‑MetRex‑8B’s 54.3\% and 7.6\% better than MasterRTL’s 65.2\%); and the overall power‑prediction pass rate advances to 70.8\% (a 7.1‑point advantage over MasterRTL’s 63.7\%). Collectively, these results confirm that our proposed regression model—combining an LLM with a MoE MLP—delivers superior accuracy for area, delay, static power and total power, substantially outperforming prior LLM‑based PPA estimators and the classical MasterRTL approach.

The proposed RocketPPA, which is fully executable on GPU, processed the entire 138 design test set in just 16s (0.12s per design) on an NVIDIA A6000 GPU. At the same time, this time for the MetRex CoT approach was $\sim300s$ on an NVIDIA H100 GPU, and $\sim514s$ for MasterRTL’s CPU preprocessing plus regression, and $\sim680s$ for a full Yosys+OpenSTA flow. Thus, the proposed LLM-based PPA estimator could provide more than 20× speedup over MetRex and 30× over MasterRTL, confirming that a direct LLM‑based Verilog to PPA regressor can surpass prior accuracy while entirely avoiding the overhead of manual feature engineering.

\begin{figure}[htbp]
  \centering
  \begin{subfigure}[b]{0.49\columnwidth}
    \centering
    \includegraphics[width=\linewidth]{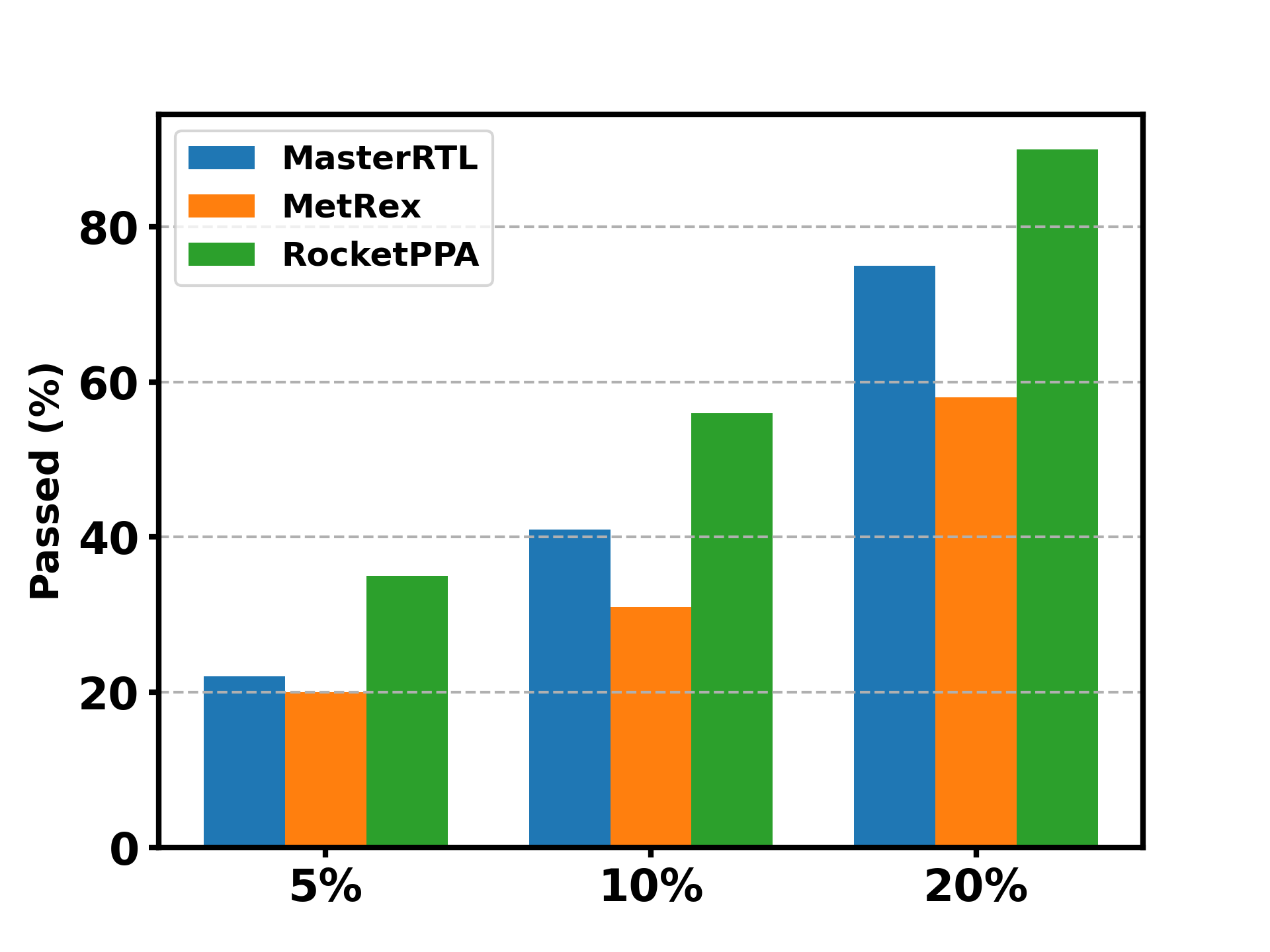}
    \caption{}
    \label{fig:l32}
  \end{subfigure}
  \hfill
  \begin{subfigure}[b]{0.49\columnwidth}
    \centering
    \includegraphics[width=\linewidth]{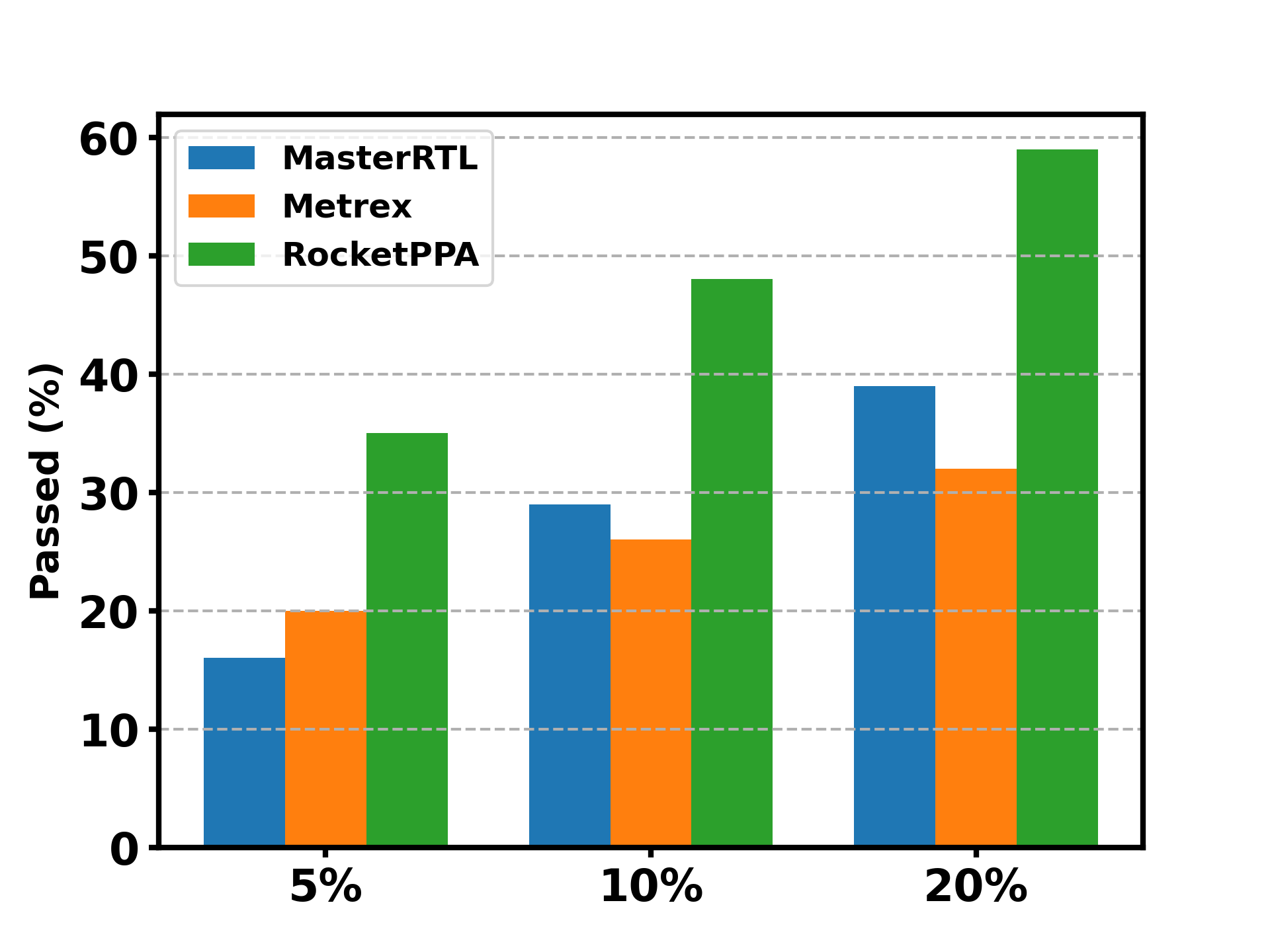}
    \caption{}
    \label{fig:l31}
  \end{subfigure}
  \caption{Comparison of pass rates for MasterRTL, Llama3‑MetRex‑8B, and PPARocket at level‑3 of the VerilogEval benchmark, with area and static power results under 5\%, 10\%, and 20\% thresholds.}
  \label{fig:l3}
\end{figure}

We conducted another comparison study by focusing on the HDL codes at Level 3 of VerilogEval, the results of which are illustrated in Figure~\ref{fig:l3}. As the results show, previous methods such as MasterRTL~\cite{ref23} and Llama3-MetRex-8B~\cite{ref11} struggle at level-3 of the VerilogEval dataset, especially under stricter thresholds for area and static power. In contrast, our model demonstrates a clear advantage in achieving the highest pass rates across all thresholds.
As an example, in the case of static power estimation, at the 10\% (20\%) threshold and MasterRTL’s pass rate decreased by 18\% (26\%), MetRex’s pass rate decreased by 16\% (22\%). In contrast, RocketPPA’s pass rate reduced by only 7\% (11\%).
Note that MetRex and MasterRTL do not report their delay estimation efficiency for the HDL codes at Level 3 of VerilogEval.

\begin{table*}
\centering
\caption{Comparison of the pass rate metric across the considered methods for Area, Delay, Static Power, and Power, evaluated under threshold values $\theta$ of 10\% and 20\% on VerilogEval benchmark.}
\label{tab:comparison}
\begin{tabular}{lcccccccc}
\toprule
& \multicolumn{4}{c}{\textbf{10\% Threshold}} & \multicolumn{4}{c}{\textbf{20\% Threshold}} \\
\cmidrule(lr){2-5}\cmidrule(lr){6-9}
\textbf{Method}
& \textbf{Area} & \textbf{Delay} & \textbf{Total Power} & \textbf{Static Power}
& \textbf{Area} & \textbf{Delay} & \textbf{Total Power} & \textbf{Static Power} \\
\midrule
\textbf{MetRex\_Llama3-8B}
& 58.0\% & 47.8\% & N/A & 42.0\%
& 75.0\% & 64.1\% & N/A & 54.3\% \\
\textbf{RocketPPA\_Llama-3.1-8B-Instruct}
& \textbf{71.6}\% & \textbf{57.2}\% & \textbf{55.0}\% & \textbf{56.7}\%
& \textbf{84.6}\% & \textbf{74.9}\% & \textbf{70.8}\% & \textbf{72.8}\% \\
\textbf{MasterRTL}
& 51.2\% & 50.7\% & 45.1\% & 47.5\%
& 78.1\% & 67.9\% & 63.7\% & 65.2\% \\
\bottomrule
\end{tabular}
\end{table*}

\subsubsection{Ablation Study}

Table~\ref{tab:combined} presents a detailed comparison of the performance of our model at 10\% and 20\%  thresholds across three design metrics (i.e., area, delay, and total power) for three foundational LLMs, each evaluated with and without the Mixture-of-Experts extension. Inclusion of the MoE architecture consistently improves accuracy in all settings. In particular, LLama-3.1-8B-Instruct with MoE achieves the highest overall performance, reaching pass rates of 71.6\%, 57.2\%, and 55.0\% at the threshold 10\%, and pass rates of 84.6\%, 74.9\%, and 70.8\% at the threshold 20\% for area, delay, and static power, respectively. 
The higher efficiency of the Llama-3.1‑8B‑Instruct variant could be because it has undergone supervised instruction tuning on a wide range of tasks, aligning its hidden‑state embeddings more closely with user‑specified objectives. In practice, this means that when we extract the final hidden layer to feed into our PPA regressor, Llama-3.1-8B-Instruct produces features that are both more semantically meaningful and more robust for downstream regression, which explains why it outperforms the Llama3.0 base model and the older CodeLlama-7B.

Furthermore, the performance gains observed when comparing the MoE variants with their respective non-MoE baselines confirm the benefit of expert specialization in enhancing prediction robustness, particularly as model size and architectural complexity increase.
We should emphasize that the cost of MoE was about 4.7M additional parameters in the case of employing LLama-3.1-8B-Instruct as the LLM engine in RockectPPA.

\begin{table*}
\centering
\caption{Ablation study on the impact of mixture-of-expert MLP across multiple base models:
Pass rates at 10\% and 20\% thresholds for Area, Delay, and total Power.}
\label{tab:combined}
\begin{tabular}{lcccccc}
\toprule
\multirow{2}{*}{\textbf{Base Model \& Method}} & \multicolumn{3}{c}{\textbf{10\% Threshold}} & \multicolumn{3}{c}{\textbf{20\% Threshold}} \\
\cmidrule(lr){2-4} \cmidrule(lr){5-7}
 & \textbf{Area} & \textbf{Delay} & \textbf{Power} & \textbf{Area} & \textbf{Delay} & \textbf{Power} \\
\midrule
\textbf{CodeLlama-7B (w/o MoE)} & 61.8\% & 47.7\% & 45.1\% & 75.8\% & 65.6\% & 57.1\% \\
\textbf{CodeLlama-7B (MoE)}    & 65.9\% & 51.7\% & 49.2\% & 79.0\% & 69.2\% & 60.2\% \\
\textbf{LLama-3.1-8B-Instruct (w/o MoE)} & 66.5\% & 53.1\% & 50.7\% & 80.9\% & 71.1\% & 63.0\% \\
\textbf{LLama-3.1-8B-Instruct (MoE)}    & 71.6\% & 57.2\% & 55.0\% & 84.6\% & 74.9\% & 70.8\% \\
\textbf{LLama3-8B (w/o MoE)} & 60.2\% & 48.1\% & 47.2\% & 73.2\% & 70.8\% & 56.9\% \\
\textbf{LLama3-8B (MoE)}    & 64.3\% & 52.1\% & 49.4\% & 76.7\% & 71.8\% & 58.1\% \\
\bottomrule
\end{tabular}
\end{table*}

\begin{figure}
  \centering
  \begin{subfigure}[b]{0.48\textwidth}
    \centering
    \includegraphics[width=0.65\textwidth]{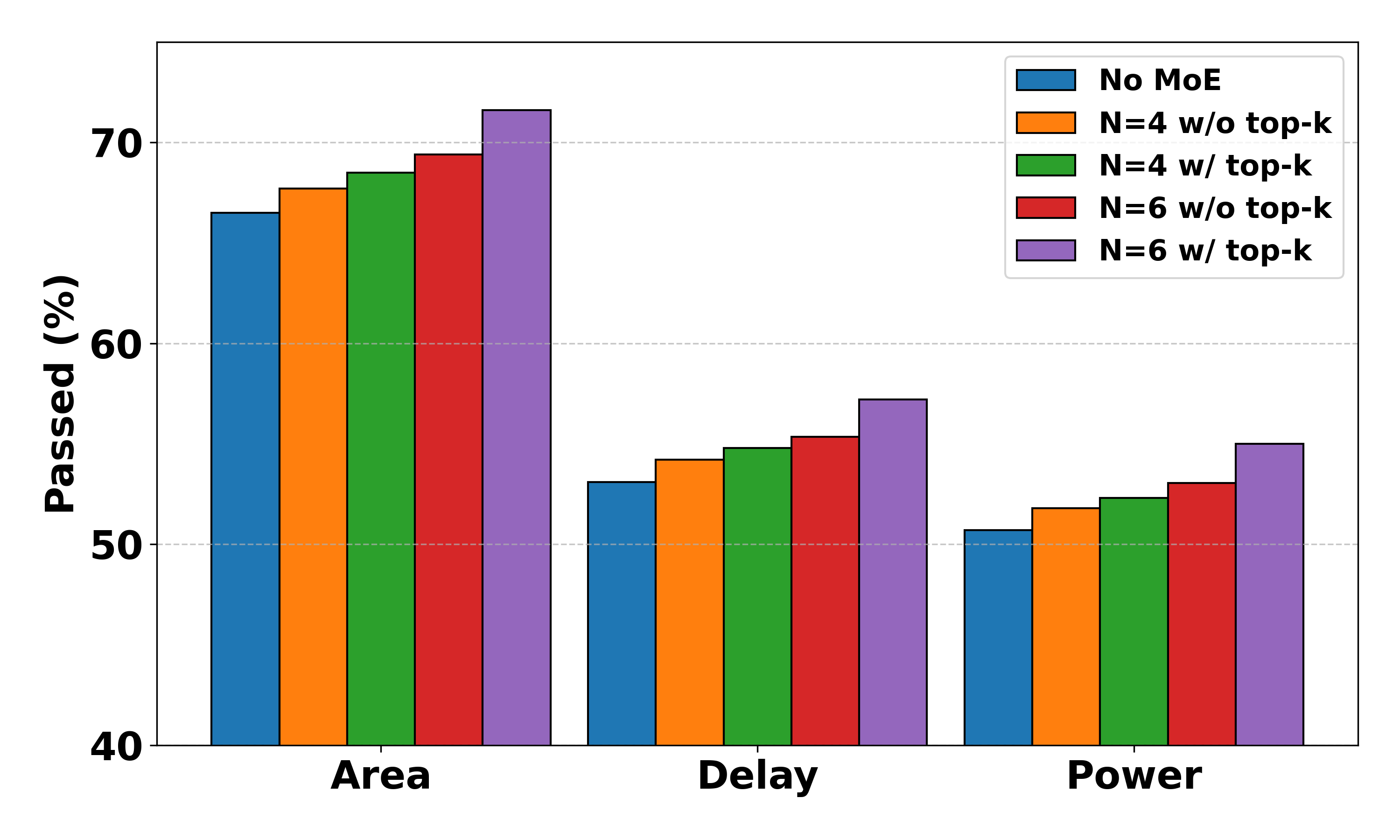}
    \caption{}
    \label{fig:comp10}
  \end{subfigure}
  \hfill
  \begin{subfigure}[b]{0.48\textwidth}
    \centering
    \includegraphics[width=0.65\textwidth]{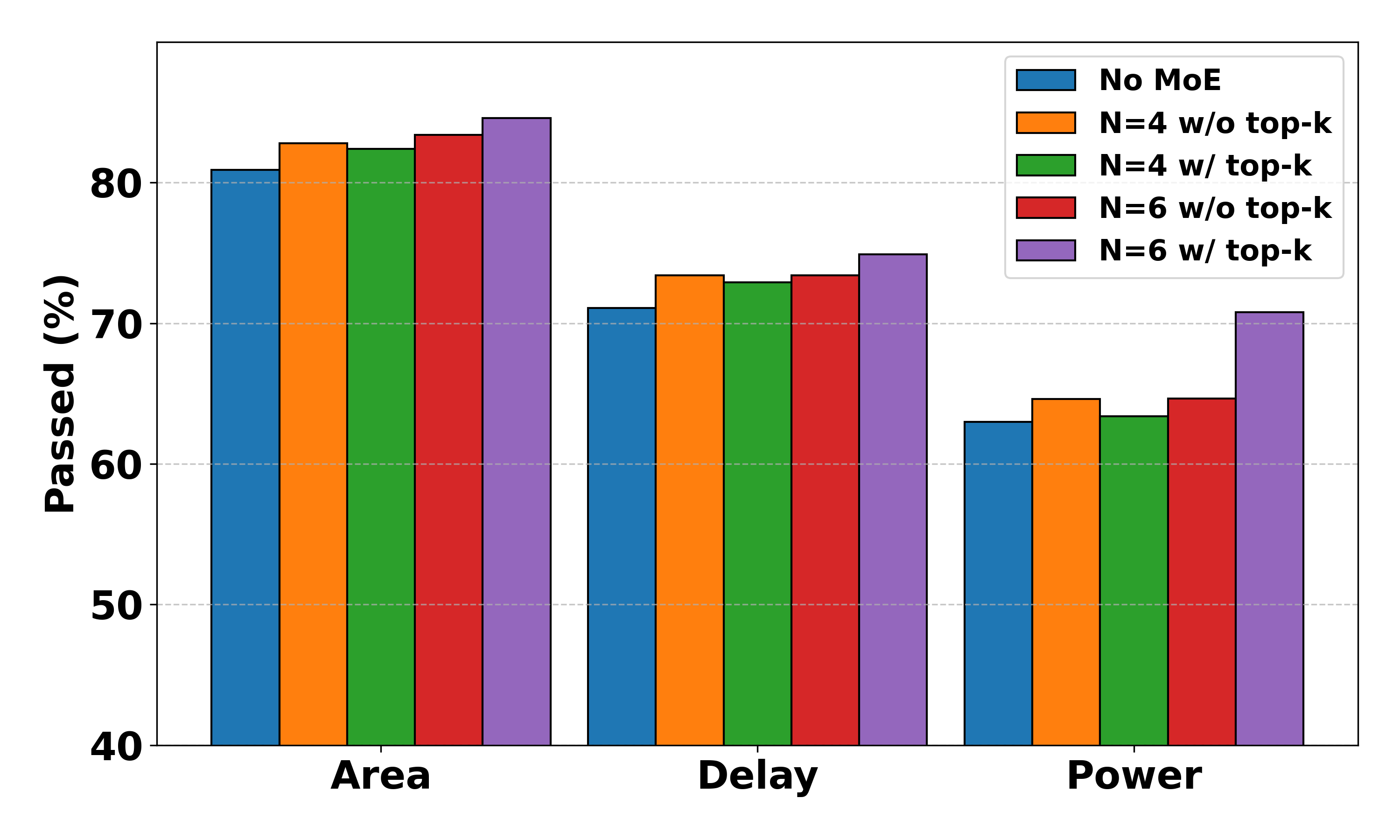}
    \caption{}
    \label{fig:comp20}
  \end{subfigure}
  
  \caption{Pass rates across different mixture-of-experts (MoE) configurations. Each bar group shows the fraction of modules with relative errors below a specified threshold for the indicated metric. }
  \label{fig:ablation_comparison}
\end{figure}

Figure~\ref{fig:ablation_comparison} shows the impact of the number of experts (parameter $N$) and the number of best selected experts (parameter $k$) used to generate the estimated value. This figure compares the pass rates at the 10\% and 20\% thresholds, respectively, when using LLama-3.1-8B-Instruc as the base model. Note that when $N$ was four (six), we considered $k$ as two (three). 
As shown, increasing the number of experts from four to six consistently increases the pass rate across all metrics (up to 3\%); adopting top‐k gating further refines the gating network’s ability to select the most relevant expert(s), yielding improved pass rates over both our single‐expert baseline and previous SOTA results. Note that our studies showed that increasing $N$ from six to a higher value leads to almost similar results, but increases the number of parameters. Therefore, six is selected as the best model.
Interestingly, the largest gains emerge in power, likely due to the higher variance in power signatures across complex modules. Moving from the 10\% threshold to the more lenient 20\% threshold raises overall pass rates by roughly 10–15\%. Still, the ranking among methods remains consistent, underscoring the general robustness of top‐k gating when combined with additional experts.

\begin{figure}
  \centering
  \begin{subfigure}[b]{0.48\textwidth}
    \centering
    \includegraphics[width=0.65\textwidth]{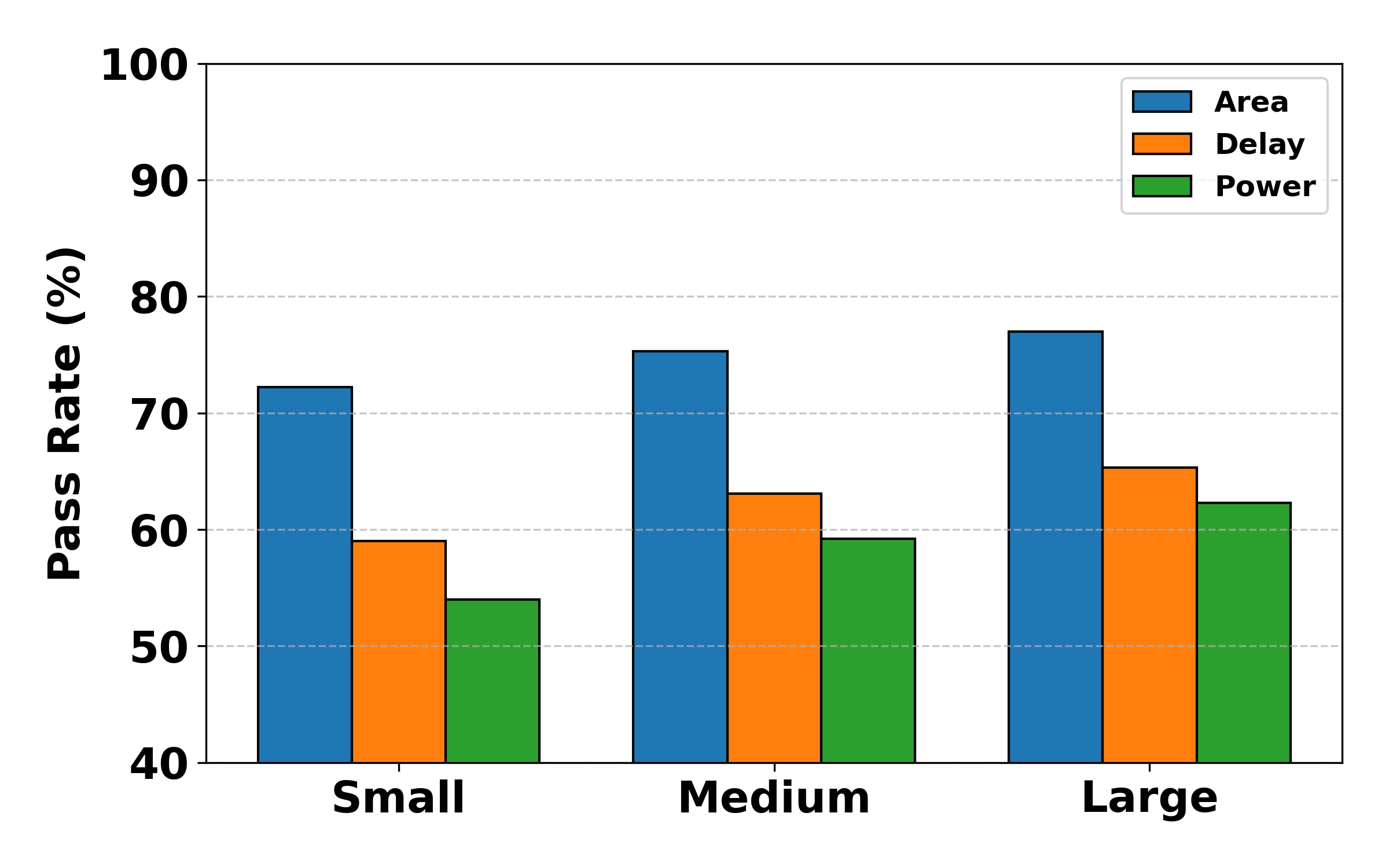}
    \caption{}
    \label{fig:val10}
  \end{subfigure}
  \hfill
  \begin{subfigure}[b]{0.48\textwidth}
    \centering
    \includegraphics[width=0.65\textwidth]{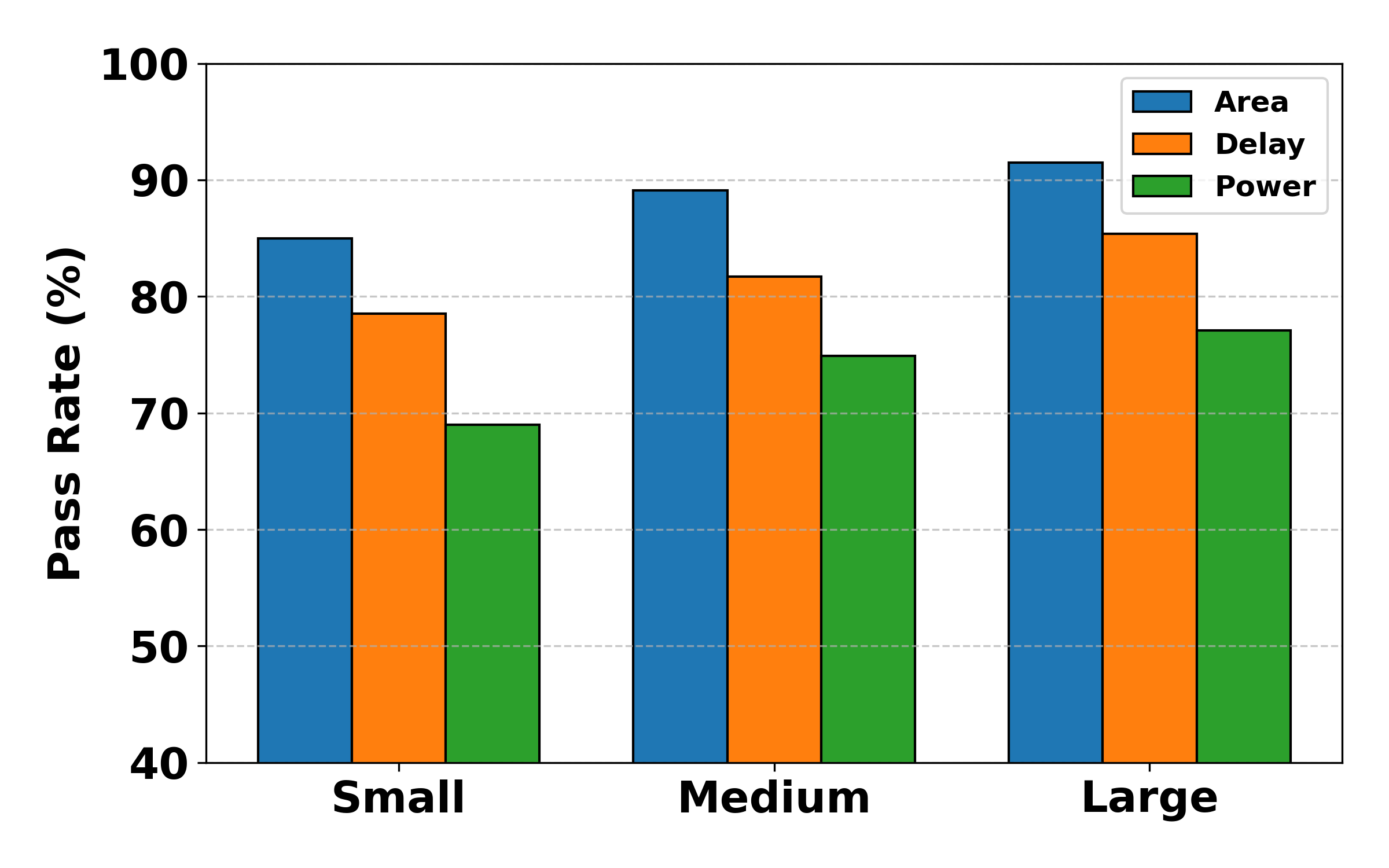}
    \caption{}
    \label{fig:val20}
  \end{subfigure}
  
  \caption{Pass Rates at 10\% and 20\% MRE Thresholds, Stratified by Token Size for LLama-3.1-8B-Instruct (MoE).}
  \label{fig:ablation_comparison2}
\end{figure}

Figure~\ref{fig:ablation_comparison2} illustrates the efficiency of the RockectPPA under different token sizes (i.e., input HDL size). The pass rates for area, delay, and total power at (a) 10\% and (b) 20\% thresholds on our generated validation set (stratified into small, medium, and large modules by token size) are reported in this figure. 
As shown, a general trend indicates that the pass rate increases with larger token sizes. For example, the pass rate of the RocketPPA for the area is increased by 5.47\% (6.52\%) for the large Verilog files compared to that of the small Verilog files under the 10\% (20\%) threshold. 
The model attains higher accuracy for larger modules, even though one might assume that more complex code would be harder to predict. A likely explanation is that lengthier Verilog files provide the model with richer contextual information (e.g., additional signals, state machines, or instantiations), enabling it to better learn and generalize power, delay, and area relationships. Meanwhile, although simpler, small modules do not offer as much structural variety, limiting the model's ability to infer PPA outcomes. 

\section{Conclusion}

We presented RocketPPA, an ultra‑fast, code‑level estimator of power, delay, and area that combines a large language model fine‑tuned with LoRA and a mixture‑of‑experts MLP regressor. Trained directly on HDL, RocketPPA surpasses state‑of‑the‑art methods—Llama3‑MetRex‑8B and MasterRTL—raising 10\%‑error pass rates by up to 13.6\% (area), 9.4\% (delay), and 14.7\% (power); even at 20\% error the gains remain 9.6–18.5\%. On the demanding Level‑3 VerilogEval set, it preserves these advantages while running more than 20x faster than MetRex and 30x faster than MasterRTL, eliminating costly feature engineering. By providing rapid, accurate PPA estimates early in the design flow, RocketPPA accelerates design‑space exploration and paves the way for next‑generation EDA tools that bridge high‑level code and physical implementation metrics.


\bibliographystyle{IEEEtran}
\bibliography{ref}

\end{document}